# Active learning in open experimental environments: selecting the right information channel(s) based on predictability in deep kernel learning

Maxim Ziatdinov,[1,2,a] Yongtao Liu,[1] and Sergei V. Kalinin[1]

[1] Center for Nanophase Materials Sciences and [2] Computational Sciences and Engineering Division, Oak Ridge National Laboratory, Oak Ridge, TN 37922, USA

Active learning methods are rapidly becoming the integral component of automated experiment workflows in imaging, materials synthesis, and computation. The distinctive aspect of many experimental scenarios is the presence of multiple information channels, including both the intrinsic modalities of the measurement system and the exogenous environment and noise signals. One of the key tasks in experimental studies is hence establishing which of these channels is predictive of the behaviors of interest. Here we explore the problem of discovery of the optimal predictive channel for structure-property relationships (in microscopy) using deep kernel learning for modality selection in an active experiment setting. We further pose that this approach can be directly applicable to similar active learning tasks in automated synthesis and the discovery of quantitative structure-activity relations in molecular systems.

[a] ziatdinovma@ornl.gov

**Introduction**

Since the earliest day of civilization, scientific research has been inseparable from explorations of similarities and correlations between observed phenomena. The examples of these range from the correlations between the phases of moon and tides, effects of temperature and annealing on mechanical properties of metals, and effects of certain plants on human health. In many cases, these correlative observations have become the foundation on which the predictive models have been built. The latter in turn lead to the understanding of associated mechanism and capability to answer the interventional and counterfactual questions, thus forming the bedrock of the modern scientific method. However, one of the most intriguing aspects of scientific discovery has been the identification of the relevant experimental parameters, suggesting the nature of the cause-and-effect relationships in the system of interest. In fact, a remarkable number of scientific discoveries – from serendipitous discovery of catalysis to the role of surface chemistry on ferroelectric behavior[1] - can be traced to the identification of the unexpected mechanism or factor affecting the experimental system.

The development of deep learning methods over the last decade has followed a very similar pattern. The availability of large volumes of imaging and text data has given an impetus to the exponential growth of supervised machine learning methods, with the deep neural networks acting as universal interpolators between large dimensional objects for problems such as regression, semantic segmentation, and translation.[2, 3] Typically the ML analyses are performed in a well-defined setting with the clearly identified feature and target data. Development of unsupervised machine learning methods exemplified by variational autoencoders allowed exploring the hidden patterns within the high-dimensional data and identifying low-dimensional mechanisms behind the observations. However, in this case, the datasets are typically well defined.

One of the examples of correlative studies is the discovery of structure-property relationships in imaging. The examples include the electron energy loss spectroscopy (EELS) in scanning transmission electron microscopy (STEM),[4, 5] relations between local structure and force-distance curves in atomic force microscopy[6] and nanoindentation,[7] and a broad variety of time- and voltage-dependent spectroscopies in piezoresponse force microscopy (PFM).[8, 9] For these methods, it has been shown that correlative relationships between local structure and functionalities can be established using the encoder-decoder-based approach, suggesting the presence of the non-observable low-dimensional latent vectors that describe the observed phenomena. While the interpretation of the latent variables in terms of relevant physical mechanisms is an open area of research, in certain cases they can be identified with order parameters or specific physical descriptors.[10, 11]

Recently, the exploration of structure-property relationships was implemented in the active learning setting.[12, 13] In this case, deep kernel learning (DKL) has been used to establish the relationship between the image patches representing local structure, and a vector representing local functionality. The DKL allows for few-shot learning, meaning that a predictive relationship can be established for a few property measurements, and is updated during the experiment once new data becomes available. The DKL-based automated experiment can be run in the classical exploration mode when total predictive uncertainty is minimized. Alternatively, a scalarizer function that converts a local functionality into a scalar measure of interest can be used to run the automated experiment. The scalarizer allows utilizing classical acquisition functions from the Bayesian optimization field, balancing the features of interest and their uncertainty to define the policy of active experiment.

However, a key but seldom explored aspect of both static and active machine learning is the capability to identify the relevant correlations from a large number of possible data channels. For example, in typical settings for supervised ML, both the target and feature sets are already defined based on domain expertise. In cases when multiple sources of data are available, if the causal relationships between the channels are known, the symbolic regression or deep learning interpolation can build the relevant structural equation models. If the causal relationships are unknown, the problem becomes that of causal discovery. Previously, we showed that the predictability and uncertainty of a physical functionality from different structural descriptors can be in principle used to probe latent physical mechanisms. However, these approaches are limited to static data sets when the full data is available for the analysis.

Here, we explore the formulation of the active experiment in the open settings, when the multiple sources of information are available and can be queried sequentially. Using the simple synthetic data set, we explore the predictability of intrinsic functionality from high-dimensional observation assuming the existence of low-dimensional hidden mechanisms. We demonstrate that in the fully Bayesian regime, the predictive uncertainty over the unmeasured data points is directly related to the predictive capacity of a channel. With this, we introduce the decrease of full predictive uncertainty as a criterion that allows channel selection in the active experiment. This approach is illustrated using multimodal imaging in piezoresponse force microscopy but can be extended to other active experimental settings including other imaging modes, materials synthesis, and theoretical calculations.

**Results and Discussion**

**I. Gaussian process and deep kernel learning.**

We start with a brief introduction to the Gaussian process and DKL. The Gaussian process (GP) model[14] with input data $X = \{x_1, x_2, ..., x_n\}$ and response (target) $y = \{y_1, y_2, ..., y_n\}$ is defined as

$$y \sim MultivariateNormal(0, K(X|\theta)) \qquad (1)$$

where $K$ is a covariate function (kernel) with hyperparameters $\theta$. Throughout this paper, we use the radial basis function (RBF) with weakly-informative priors as the kernel,

$$K_{ij} = \alpha \exp\left(-0.5 \left\|x_i - x_j\right\|^2 / l^2\right) + \delta_{ij}\sigma_{noise}, \qquad (2a)$$

$$\alpha \sim LogNormal(0, 1), \qquad (2b)$$

$$l \sim LogNormal(0, 1), \qquad (2c)$$

where $\sigma_{noise}$ corresponds to a noise term in the regression setting. The GP posterior is usually derived analytically and then directly sampled from. Specifically, the posterior predictive distribution at new inputs $X_*$ is defined as

$$p(y_*|X_*, y, X) = \mathcal{N}(\mu_*, \Sigma_*), \qquad (3a)$$

$$\mu_* = K(X_*, X|\theta)K(X|\theta)^{-1}y, \qquad (3b)$$

$$\Sigma_* = K(X_*|\theta) - K(X_*, X|\theta)K(X|\theta)^{-1}K(X, X_*|\theta), \qquad (3c)$$

where $K(X|\theta)$ and $K(X_*|\theta)$ correspond to the kernel computation between all pairs of original input data ('training data') and all pairs of new input data ('test data'), respectively, and $K(X_*, X|\theta)$ is the matrix of covariances between the original and new inputs. The kernel hyperparameters are usually inferred from the training data using either Markov chain Monte Carlo sampling or variational inference approaches. The equations (3) thus comport to the classical Bayesian learning, i.e., update of the non-parametric model parameters given new experimental data.

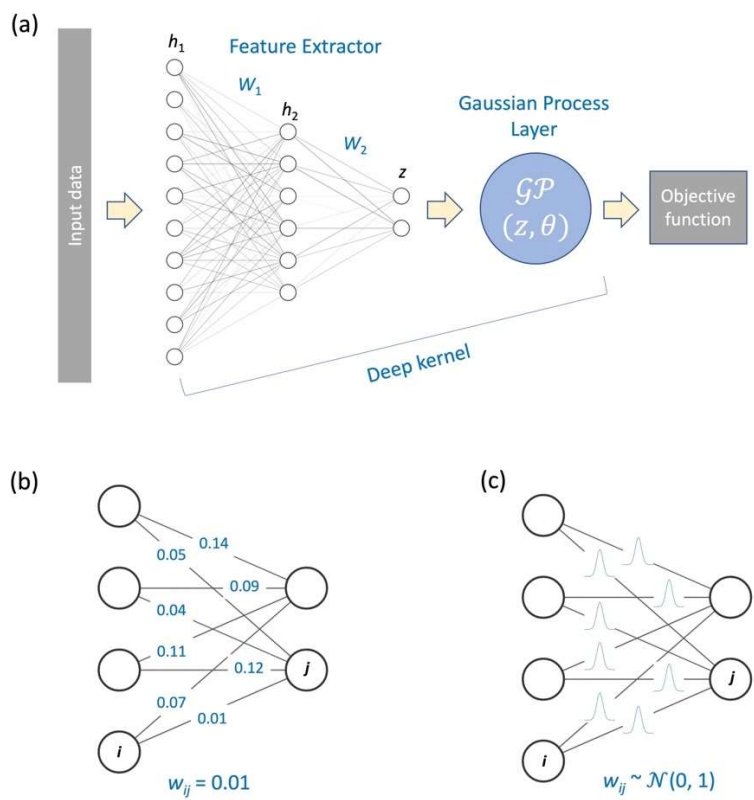

**Figure 1.** (a) Schematics of deep kernel learning. A neural network ("Feature Extractor") embeds high-dimensional data into the latent space *z* where a standard Gaussian process kernel operates. (b) In the deterministic neural networks, each weight is represented by a single value. (c) By contrast, in the fully Bayesian neural networks, each weight is represented by a prior probability distribution, here chosen as a standard normal distribution. During the training, the probability distributions for individual weights are updated, analogous to scalar weight values for classical neural networks.

The main limitation of GP with most standard kernels is that it does not learn representations from data, which precludes it from working with high-dimensional data and analyzing relationships between different data modalities (e.g., structure-property relationships in imaging and spectroscopy). The deep kernel learning (DKL) approach, originally introduced by A.G. Wilson *et al.*,[15] tries to address this limitation by combining GP with a (deep) neural network (Fig. 1a). The deep neural networks are known for their ability to learn powerful representations from high-dimensional data that can aid in predictions on new, previously unseen by the network, inputs (as long as they come from the same distribution as the training data). Computation-wise, in DKL, a neural network embeds the high-dimensional input data into the low-dimensional 'latent space' where a standard GP kernel operates. The weights of the neural network and the parameters of the GP base kernel are learned jointly from the training data. Hence, this structure (neural network + GP) can be referred to as 'deep kernel' and it can be used as a drop-in replacement of the regular GP kernel in Eq 1 and 3:

$$K_{DKL}(X|\omega, \theta) = K_{base}(g(X|\omega)|\theta) \quad (4)$$

where $g$ denotes a neural network with weights $\omega$.

The 'deep kernel' hyperparameters are usually optimized with respect to marginal likelihood via (stochastic) gradient descent. This approach is sometimes referred to as 'empirical Bayesian'. It is computationally cheap but has a disadvantage because it tends to overfit the data resulting in an unreliable performance on new inputs.[16] Another way to learn 'deep kernel' hyperparameters is via Markov chain Monte Carlo (MCMC) sampling techniques. In this case, the deterministic weights of the neural networks are replaced by probabilistic distributions (Fig. 1b, c), and the MCMC is used to obtain posterior samples for both neural network weights and base kernel hyperparameters. Finally, a somewhat intermediate approach is using an ensemble of DKL models with different random initialization of neural network weights. The ensembles of deep neural networks have shown a robust performance for multiple deep learning problems, both in terms of prediction accuracy and uncertainty estimates, and are sometimes viewed as a potential substitute for the fully Bayesian treatment.[17, 18]

**II. Latent space of DKL: toy example**

As a model case, here we assume that the state of the system is described by the (non-observable) vector of structural parameters, $z$. Experimentally measured is the scalar physical functionality, $y$. We assume that $y = f(z)$, where $f$ is some non-linear function. We further assume that $z$ is not observed, but we have access to high-dimensional noisy observations, $X$, that contain partial information about $z$. It is important to note that the relationships between $X$ and $z$ is noisy, $X$ is also affected by other non-observed variables, and may depend only on the subset of $z$. First, we explore the potential of DKL to predict $y$ from $X$ and explore a relation between embedded variables of the DKL model and 'ground truth' parameters $z$.

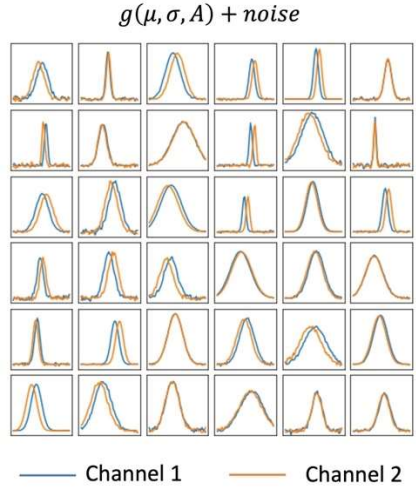

**Figure 2.** Random samples from the single peak dataset with varying peak position, peak width, peak amplitude, and noise level. The data in Channel 2 is produced by shifting the peak position in Channel 1 by a small $\Delta$ sampled from a uniform distribution.

We generate our high-dimensional input data from three uniformly distributed 'latent' variables, $\mu, \sigma, A$, using a Gaussian function, $g(\mu, \sigma, A)$, and adding white noise to it. Hence, our dataset consists of noisy1D peaks with different amplitudes, centers, and widths (Figure 2). Next, we generate our target variable as $y = 0.5\mu^3 + 4\sigma^2$. In other words, we aim to discover the $y$ given the observations $X$ of the curves in Fig. 2, without the knowledge of the functional form of $y$ on the ground truth parameters. Note that the target variable depends only on the first two latent variables, but not on the third one.

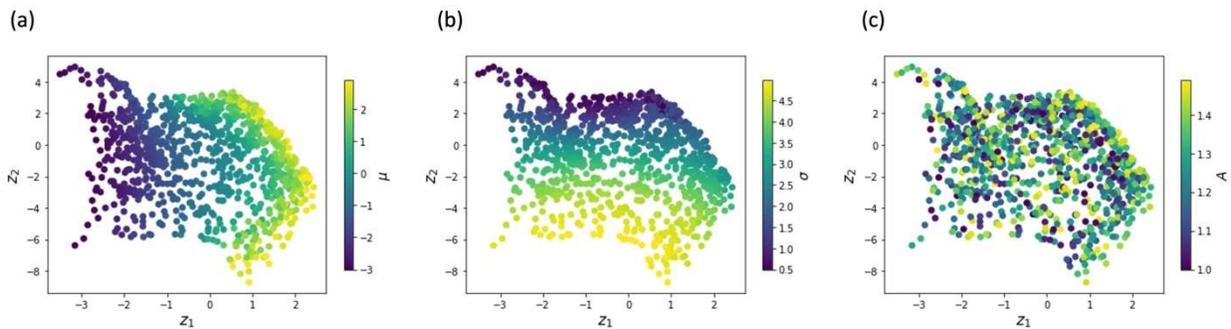

**Figure 3.** Latent space of a single DKL model with the entire peak 'spectra' (see Fig. 2) as inputs and target values corresponding to $y = 0.5\mu^3 + 4\sigma^2$. The colormap corresponds to the ground truth values of peak center (a), peak width (b), and peak amplitude (c). Note that the trends in the latent space are visible even though $y$ is a non-linear function of $\mu$ and $\sigma$.

We train DKL on the entire dataset of 1000 curves and use the neural network of the trained model to find a 2D embedding of those curves. The results are shown in Fig. 3 and demonstrate that the first latent dimension ($z_1$) is mainly associated with the peak position $\mu$ whereas the second latent dimension ($z_2$) is mainly associated with the peak width $\sigma$. At the same time, we see that the amplitude $A$ was marginalized out since the response variable does not depend on it. We note that while for the toy problems, the interpretation of the latent variables can be straightforward, this may not always be the case for more complex systems. Furthermore, the DKL setup (similar to conventional autoencoders) does not explicitly encourage the full disentanglement of the factors of variation, and the full disentanglement may not always be necessary to obtain a high accuracy prediction for a target variable.

**III. Channel selection based on predictive uncertainty**

We next illustrate how one can utilize DKL predictive uncertainty to select the channel with the best predictive capacity, i.e., the channel that allows for the most accurate reconstruction of the target function. To do this, we create a second channel (Channel 2) by shifting the peak position in Channel 1 by a small $\Delta$ sampled from a uniform distribution (see Fig. 2). We then train DKL on a small subset of data and compare the predictive uncertainties on the remaining data. We expect that the correct channel will produce a lower mean/total uncertainty. We introduce an 'accuracy' metric which indicates how many times the correct channel had lower predictive uncertainty out of 30 different random initializations of training data. Finally, we compare the accuracy of a single-model DKL optimized via gradient ascent on the log marginal likelihood, the ensemble of DKL models, and the fully Bayesian DKL model whose parameters are inferred via Hamiltonian Monte Carlo (HMC).

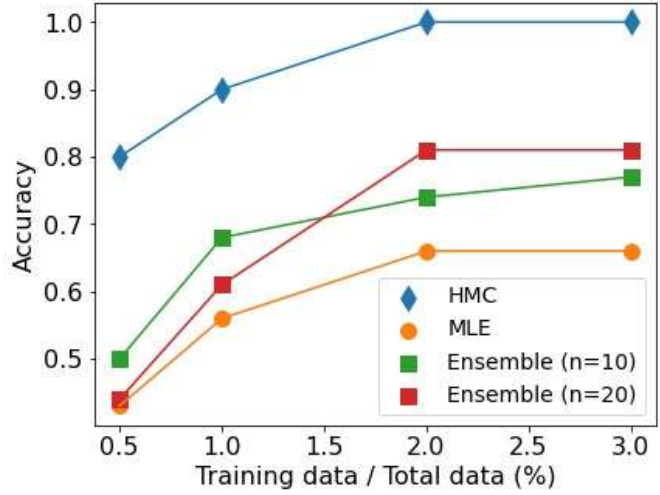

**Figure 4.** Identification of the channel with best predictive capacity via deep kernel learning with Hamiltonian Monte Carlo (HMC), gradient ascent on log marginal likelihood (MLE), and ensemble of *n* MLE models with different random initialization of their parameters. The horizontal axis shows the percentage of data used for model training. The vertical axis shows the number of correct identifications for 30 different initializations of training data (the maximum/minimum value is one/zero).

The results in Figure 4 show that the HMC-DKL has the most robust performance and outperforms both single model and ensemble DKL, especially in the small data regime. The ensemble DKL generally shows a better performance than a single model, and, in principle, can serve as an approximation for HMC-DKL for the purpose of channel selection in situations where performing fully Bayesian inference is computationally not feasible.

**IV. Information channel discovery in the active learning setting: toy data**

Having confirmed on the static snapshots of data that predictive uncertainty can be used to identify channel with the best predictive capacity, we proceed to the active channel learning where the goal is to learn the correct channel starting from a small number of measurements and use it to guide the selection of the next measurement points, meaning the channel to sample and region of parameter space within the channel. This active channel learning procedure is summarized in the Algorithm 1.

---
**Algorithm 1 (Active Channel Learning)**

**Inputs:** Initial training set, $D = \{x^i, y^i\}_{i=1,...,N}$; Array of unmeasured points $X_*$; Acquisition function $\alpha$; Reward function $R$; Channel rewards $R_a$ (initialized at zeros); $\varepsilon$ in $\varepsilon$-greedy policy; Number of warm-up and exploration steps, $N_{\text{warmup}}$ and $N_{\text{explore}}$

**for** $i=1, \ldots, N_{\text{warmup}}$ **do**
   Train independent DKL model for each channel (in parallel or sequentially)
   Compute posterior predictive uncertainty, $\mathbb{V}[y_*]$, over $X_*$, for each channel
   Reward the channel with the lowest mean uncertainty, $V_m = \frac{1}{N}\sum_{i=1}^{N} \mathbb{V}[y_*^i]$
   Use the rewarded channel to derive acquisition function, $\alpha(\mathbb{E}(y_*), \mathbb{V}[y_*])$
   Perform the next measurement in $x_{\text{next}} = argmax(\alpha)$. Update $D$ and $X_*$
**end**
Average channel rewards over the warmup steps, update $R_a$

**for** $i=1, \ldots, N_{\text{steps}}$ **do**
   Use $\varepsilon$-greedy policy to sample a channel number
   Train DKL model using the selected channel
   Compute posterior predictive uncertainty, $\mathbb{V}[y_*]$, over $X_*$
   Reward/penalize according to $R$ ($\mathbb{V}_m^i$, $\mathbb{V}_m^{i-1}$) and update $R_a$
   Compute $\alpha(\mathbb{E}(y_*), \mathbb{V}[y_*])$ and find $x_{\text{next}} = argmax(\alpha)$
   Perform measurement in $x_{\text{next}}$. Update $D$ and $X_*$
**end**

---

We start by creating the initial dataset where for a small number of randomly sampled high-dimensional multi-channel inputs there are corresponding scalar targets. We then proceed to the warm-up phase during which a separate DKL model is trained for each channel. The channel associated with the DKL model that produced the lowest mean predictive uncertainty on the unmeasured data receives a positive reward. The rewarded model is also used to derive the next measurement point based on the pre-defined acquisition function. Here, our acquisition function simply selects the point with the largest value of uncertainty in the DKL prediction. The warm-up phase is necessary because the correct channel identification is less robust in the small data regime (see Figure 4). After the competition of the warm-up phase, we sample a single channel at each active learning step based on the $\varepsilon$-greedy policy and use it to obtain a posterior predictive distribution from which we derive the acquisition function for selecting the next measurement point.

Figure 5 shows the results of applying Algorithm 1 to the 2-channel dataset with 1D peaks (see Figure 2) for the ensemble-DKL (Fig. 5a) and HMC-DKL (Fig. 5b). In both cases, we were able to converge on the correct channel. At the same time, the evolution of the mean predictive uncertainty as a function of the active learning step (including warmup steps) appears to be noisier in the case of ensemble-DKL. The final sample-averaged rewards are 0.455 (Channel 1) and -0.125 (Channel 2) for the ensemble-DKL and 1.0 (Channel 1) and -0.125 (Channel 2) for the HMC-DKL.

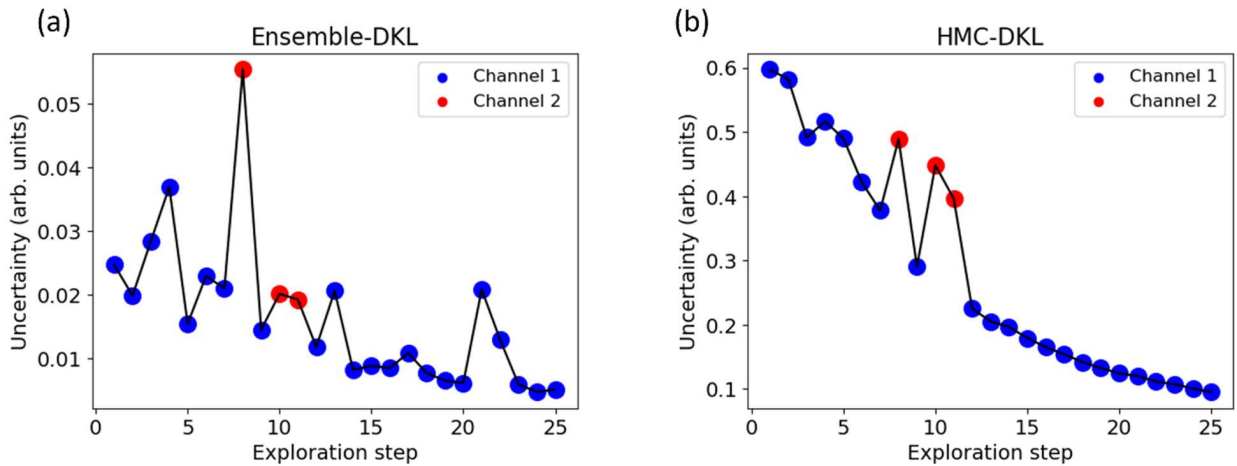

**Figure 5.** The evolution of the mean predictive uncertainty as a function of the number of active learning steps for ensemble-DKL (a) and HMC-DKL (b). In both cases, we start with 2% of the data and use 5 warm-up states during which both channels are evaluated in parallel and the one with the lowest uncertainty is used for selecting the next evaluation point. After the warmup phase, we sample a single channel at each step based on the epsilon-greedy policy with epsilon decreased uniformly ("annealed") from 0.4 to 0.1 during the 20 steps. The data in both cases was standardized by subtracting mean and dividing by standard deviation.

## V. Application for experimental data

Finally, we apply active channel learning to real experimental data in multimodal imaging. As a model system, we use a pre-acquired high grid density band excitation Piezoresponse spectroscopy (BEPS) data set of a PbTiO$_3$ (PTO) thin film (see Methods for more details about the PTO sample). The principles of the PFM and BEPS are discussed elsewhere.

This data set contains four image channels (left hand side of Figure 6) showing spatial variability of material's properties. Here, the amplitude channel represents the magnitude of the electromechanical response and delineates ferroelectric and ferroelastic domain patterns. The phase channel shows the ferroelectric polarization orientation. The frequency and quality factor channels depict the resonance frequency and quality factor of the cantilever in contact with the surface. The magnitudes of corresponding signals are related to local elastic property and energy dissipations at the tip-surface junction, respectively. In addition to four image channels, the data set also contains spectroscopy data (right hand side of Figure 6), e.g., polarization-voltage, and frequency-voltage hysteresis loops, at each location, which represent the evolution of materials response as a function of voltage. Note that this data set is used in our previous work,[13] here we only reuse the data to demonstrate the application of active channel learning.

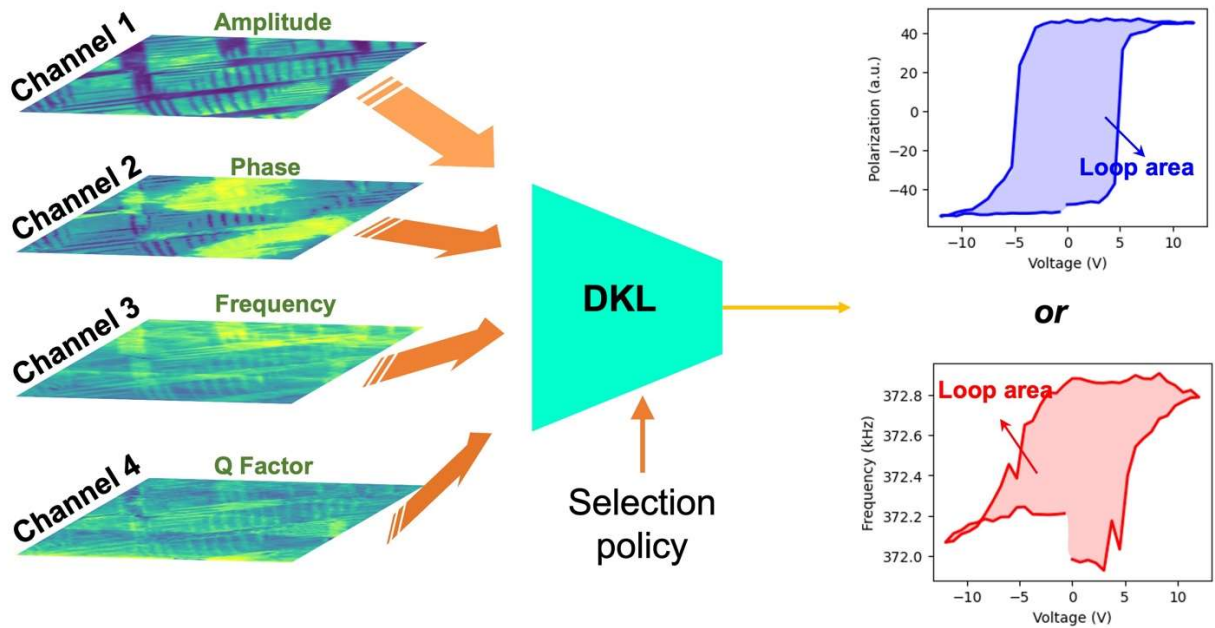

**Figure 6.** Experimental multimodal piezoresponse force microscopy data from PbTiO$_3$ film used to predict a functional property. The latter is represented by the polarization loop areas. The structural data has four channels: band excitation amplitude (Channel 1), phase (Channel 2), resonance frequency (Channel 3), and Q-factor (Channel 4). The goal is to identify the channel with the best predictive capacity for the hysteresis loop area.

The approach of analyzing this BEPS data set is outlined in the Algorithm 2. Its main part is the same as in Algorithm 1 but unlike the case of synthetic data, it requires several additional steps involving data acquisition and data pre-processing. In particular, it requires a selection of physics-based 'scalarizer' that converts the spectroscopic measurement to a scalar value associated with the physical functionality of interest and represents the degree of interest in a specific aspect of the data. Here, we used the polarization-voltage loop area and frequency-voltage loop area as the measures of the energy loss during switching and voltage-induced irreversible dynamics, respectively.

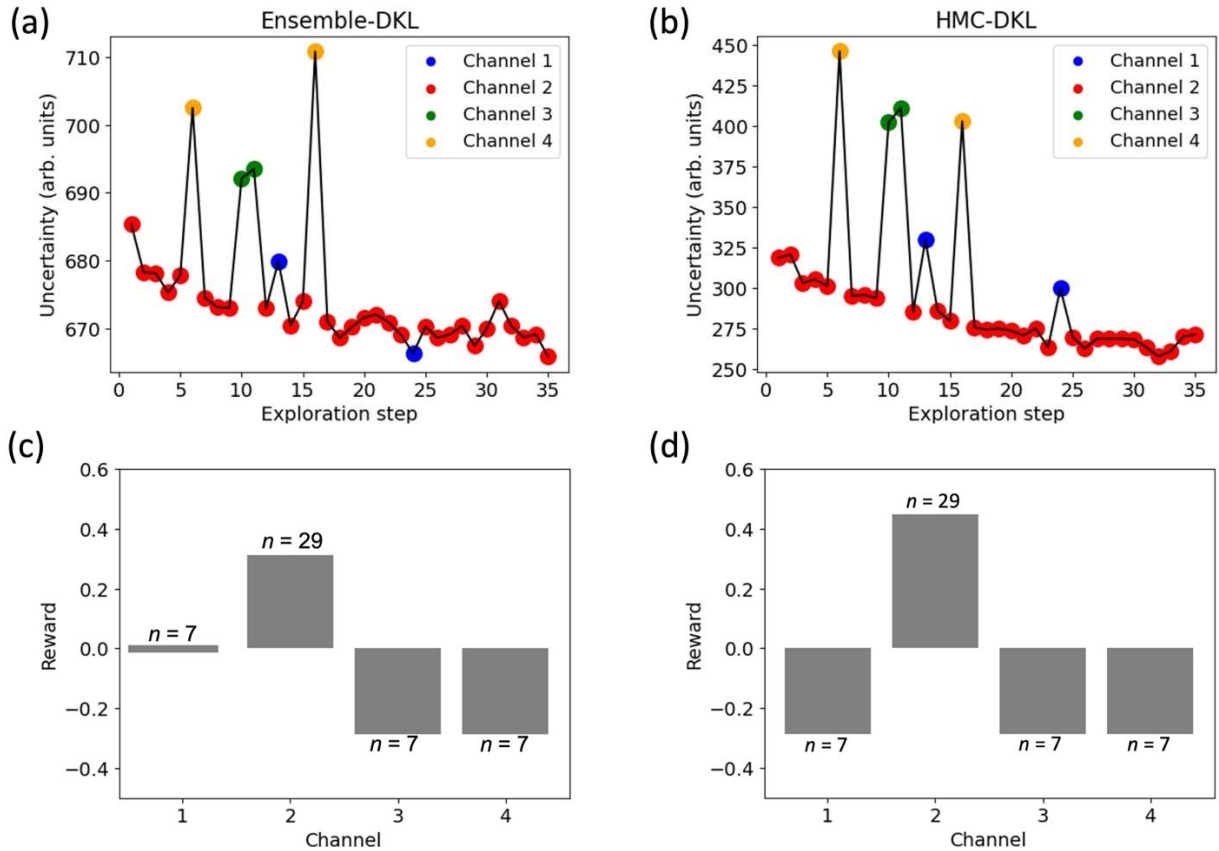

**Figure 7.** Multichannel DKL analyses of BEPS polarization-voltage loop area. The evolution of the mean predictive uncertainty as a function of active learning steps for ensemble-DKL (a) and HMC-DKL (b). In both cases, we start with 1% of the data and use 5 warm-up states during which all the channels are evaluated in parallel and the one with the lowest uncertainty is used for selecting the next evaluation point. After the warmup phase, we sample a single channel at each step based on the epsilon-greedy policy with epsilon decreased uniformly ("annealed") from 0.4 to 0.1 during the 30 steps. (c, d) The final sample-averaged rewards obtained by each channel for Ensemble-DKL (c) and HMC-DKL (d). The *n* corresponds to the number of times a particular channel was sampled (including warmup steps). The input and target data in both cases were normalized to (0, 1).

The results of active channel learning of polarization-voltage loop area are shown in Figure 7. We found that both ensemble-DKL and HMC-DKL tend to converge on the second channel corresponding to the band excitation phase (Fig. 7a, b). In agreement with the results obtained on synthetic data, the HMC-DKL demonstrates a smoother performance and allows for a better delineation of the channels (Fig. 7c, d).

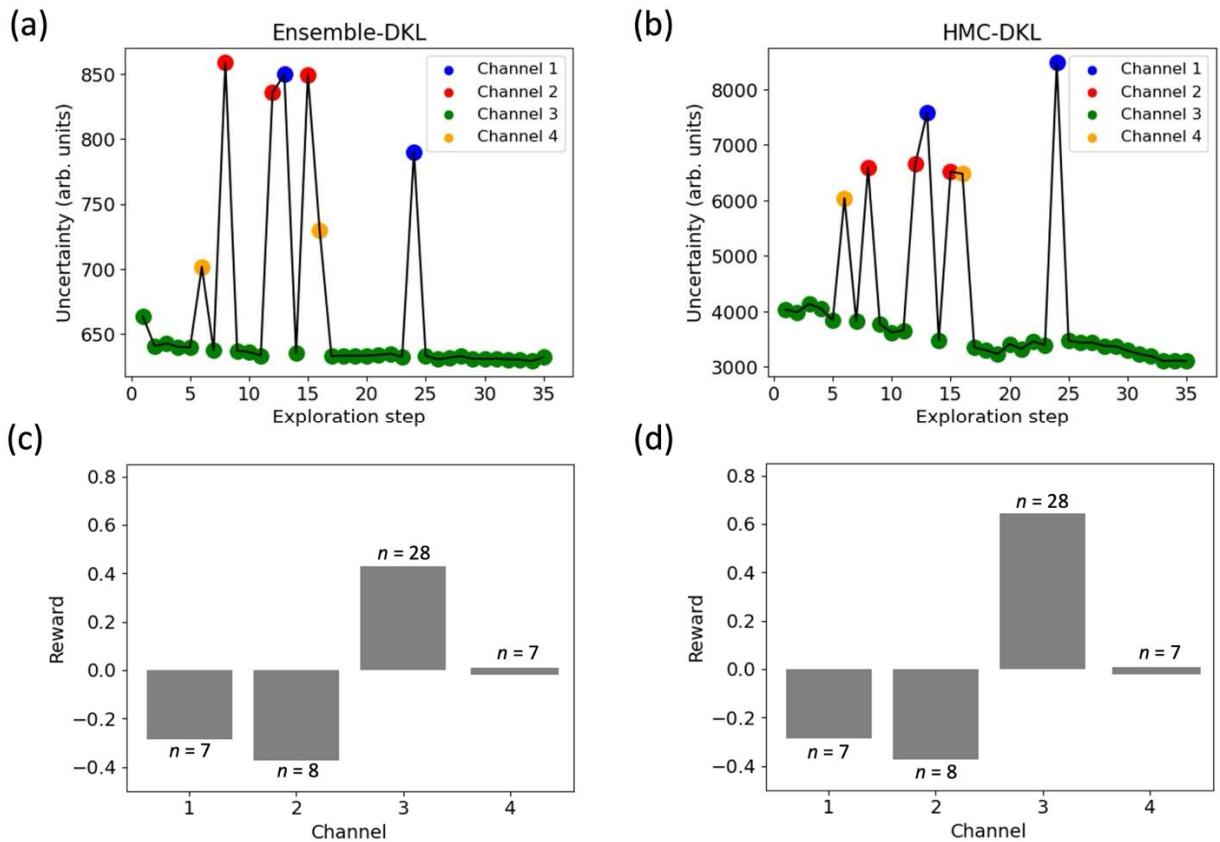

**Figure 8.** Analyses of BEPS frequency-voltage loop area. The evolution of the mean predictive uncertainty as a function of active learning steps for ensemble-DKL (a) and HMC-DKL (b). In both cases, we start with 1% of the data and use 5 warm-up states during which all the channels are evaluated in parallel and the one with the lowest uncertainty is used for selecting the next evaluation point. After the warmup phase, we sample a single channel at each step based on the epsilon-greedy policy with epsilon decreased uniformly ("annealed") from 0.4 to 0.1 during the 30 steps. (c, d) The final sample-averaged rewards obtained by each channel for Ensemble-DKL (c) and HMC-DKL (d). The $n$ corresponds to the number of times a particular channel was sampled (including warmup steps). The input data in both cases was normalized to (0, 1) whereas the target data in both cases was standardized by subtracting mean and dividing by standard deviation.

In comparison, we also perform the analyses of frequency-voltage loop area, the results of are shown in Figure 8. In this case, both ensemble-DKL and HMC-DKL converge on the third channel corresponding to frequency (Fig. 8a, b). Both analyses of polarization-voltage loop area and frequency-voltage loop area show reasonable results, where polarization-voltage loop area converges on a channel related to polarization property (i.e., phase channel represents polarization orientation) and frequency-voltage loop area converges on a channel related to frequency.

---

**Algorithm 2 (Active Channel Learning for Structure-Property Relationships)**
**Inputs:** Full multi-channel structural image $X$ of size $m \times n$; Patch size $p$; Physics-based scalarizer $f_s$ for functional measurements; Reward function $R$; Channel rewards $R_a$ (initialized at zeros); $\varepsilon$ in $\varepsilon$-greedy policy; Number of warm-up and exploration steps, $N_{warmup}$ and $N_{steps}$

Create a stack of image patches $x_{k,l}$ of size $p \times p$ from $X$, where $(k, l)$ are grid indices
Perform $N$ spectroscopic measurements at random points on the grid ($N \ll m * n$)
Split image patches into the measured, $X_{meas}$, and unmeasured, $X_*$
Create initial training set $D = \{x^i_{meas}, f_s(y^i)\}_{i=1,\dots,N}$
Run **Algorithm 1**

---

**Conclusions**

To summarize, here we propose the approach for the active learning from multiple information channels towards the physical discovery. In this approach, the algorithms has access to multiple information channels in active regime, and aims to explore the channels towards the highest predictability of functionality of interest. Here, we implemented this approach for the model systems and multimodal piezoresponse force microscopy, for which we establish which imaging channel allows best predictability of the spectroscopic functionality.

This approach can be extended in a straightforward manner to other imaging techniques, including multimodal electron microscopy, other scanning probe microscopy techniques, and other forms of chemical and optical imaging targeting the discovery of specific functionalities of interest. Note that in these cases one can explore the intrinsic measurement channels of the imaging tool, as well as environmental signals such as humidity, noise levels, etc.

Furthermore, this analysis allows for generalization for other experimental settings, including chemical synthesis and theoretical prediction. For example, for synthesis the typical application will be building the relationship between compositions and processing conditions and functionalities of interest. For theoretical modelling, this includes discovering optimal predictor for the specific functionality of interest.

Finally, we note that this work sets the foundation for the optimal channel discovery in the active learning setting. The further extension of this work will include generalization to the discovery of predictability from the linear combinations of input channels and more complex functions.

**Methods**

*DKL*

We used a 3-layer MLP with the hyperbolic tangent activation function in all DKL models. For the MLE-based single and ensemble DKL, the Adam optimizer[19] with the learning rate of 0.001 was used. In the HMC-DKL, we used the iterative No-U-Turn sampler[20] with 2000 warm-up steps and 2000 samples for the toy data, and with 1000 warm-up steps and 1000 samples split into 3 parallel chains for the PFM data. The detailed methodologies of active channel learning are available via Jupyter notebooks at https://github.com/ziatdinovmax/ActiveChannelLearning.

*PTO sample*

The PTO thin film with a $SrRuO_3$ conducting buffer layer was grown by metalorganic chemical vapor deposition (MOCVD) method on (001) $KTaO_3$ substrates. The material growth was previously reported by H. Morioka et al.[21]

*PFM BEPS measurements*

The PFM BEPS data were acquired with a National Instruments DAQ card and chassis operated with a LabView framework using an Oxford Instrument Asylum Research Cypher microscope with Budget Sensor Multi75E-G Cr/Pt coated AFM probes (~3 N/m).

**Conflict of Interest**

The authors declare no conflict of interest.

**Authors Contribution**

M.Z. and S.V.K. conceived the project. M.Z. designed and implemented algorithms for active channel learning. Y.L. collected PFM data and assisted in its analysis. M.Z. and S.V.K. wrote the manuscript with inputs from Y.L.

**Acknowledgments**

The machine learning effort was supported by the Oak Ridge National Laboratory's Center for Nanophase Materials Sciences (CNMS), a U.S. Department of Energy, Office of Science User Facility (M.Z.). The SPM measurements were supported by the center for 3D Ferroelectric Microelectronics (3DFeM), an Energy Frontier Research Center funded by the U.S. Department of Energy (DOE), Office of Science, Basic Energy Sciences under Award Number DE-SC0021118 (Y.L., S.V.K.).

**Data Availability Statement**

The data that support the findings of this study are available at https://github.com/ziatdinovmax/ActiveChannelLearning